# Human-Robot Cooperative Piano Playing with Learning-Based Real-Time Music Accompaniment

Huijiang Wang, *Student Member, IEEE,* Xiaoping Zhang, *Member, IEEE,*
Fumiya Iida, *Senior Member, IEEE*

*Abstract*—Recent advances in machine learning have paved the way for the development of musical and entertainment robots. However, human-robot cooperative instrument playing remains a challenge, particularly due to the intricate motor coordination and temporal synchronization. In this paper, we propose a theoretical framework for human-robot cooperative piano playing based on non-verbal cues. First, we present a music improvisation model that employs a recurrent neural network (RNN) to predict appropriate chord progressions based on the human's melodic input. Second, we propose a behavior-adaptive controller to facilitate seamless temporal synchronization, allowing the cobot to generate harmonious acoustics. The collaboration takes into account the bidirectional information flow between the human and robot. We have developed an entropy-based system to assess the quality of cooperation by analyzing the impact of different communication modalities during human-robot collaboration. Experiments demonstrate that our RNN-based improvisation can achieve a 93% accuracy rate. Meanwhile, with the MPC adaptive controller, the robot could respond to the human teammate in homophony performances with real-time accompaniment. Our designed framework has been validated to be effective in allowing humans and robots to work collaboratively in the artistic piano-playing task.

## I. INTRODUCTION

Human-robot cooperation (HRC), which characterizes the scenario where humans and robots collaborate in their environment to develop a tightly coupled system that achieves a common objective[1], has long been a topic of interest in robotics research. In the human-robot cooperation, the ability of robots to work alongside humans has facilitated numerous applications, ranging from manufacturing and logistics[2] to healthcare and entertainment[3], [4], [5]. At the same time, thanks to the advances in physical human-robot interaction (pHRI) technology, information interaction between human and robots becomes possible, which further promotes the realization of human-robot cooperation [6], [7], [8], [9].

The human-robot collaborative manipulation proposes that humans and robots work together in a shared workspace to achieve a shared goal, where the robot is to appropriately respond to the actions and behaviors of the human operator and compensate for the human clumsiness or mistakes[10], [11], [12]. The robots in human-robot cooperation, named as cobots, have already been widely adopted, from simple repetitive actions such as assembly task[13], [14], [15], [16], pick-and-place task[17] to complex operations that require a high level of dexterity and skill including healthcare[18], monitoring and rescue tasks[19], [20], and service tasks[21]. However, the use of cobots in creative tasks, such as playing musical instruments, is still a relatively unexplored area of research. In the field of music, robots have shown great potential for enhancing musical creativity, performance, and education. However, achieving successful human-robot collaboration in musical contexts is still a challenging task.

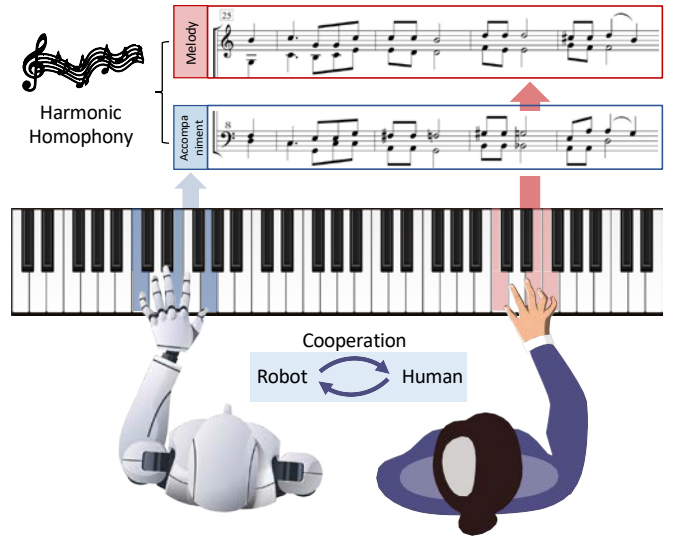

Fig. 1. The system overview of the human-robot cooperative piano playing. The robot co-player is assisting a human pianist by providing harmonic chord accompaniment for the human-played melody by understanding the non-verbal cues of its human teammate, which ultimately produces a homophony performance. The musical notation indicates the temporal synchronization between the melodic line (upper, red) and the harmonic accompaniment (bottom, blue) that supports it.

Playing the piano involves intricate rhythmic motions that demand precise coordination of the upper limbs. Much of the existing research on piano robots centers around physically reproducing human keystrokes. Various studies have explored the development of human-like hands and upper limbs for implementing striking on piano notes including anthropomorphic skeletons [22], [23], variable stiffness hands [24], [25], [26], and electroactive artificial muscles [27]. Additionally, attention has been given to the control and algorithm design of piano-playing robots to achieve smooth note arrangements for enhanced musical performance [28], [29], [30], [31], [32]. More

The authors are with the Bio-Inspired Robotics Lab, Department of Engineering, University of Cambridge, Cambridge CB2 1PZ, U.K. Corresponding Author: Huijiang Wang, E-mail: hw567@cam.ac.uk.

recently, researchers have focused on replicating humanoid piano performances by considering muscle activities [33], [34], [35] in virtual simulators and models. Despite numerous studies primarily concentrating on imitating keypress motions on robot platforms, there is still room for exploration into the creation of artistic and harmonious musicality.

Executing a piano duet demands not only the artful arrangement of notes, exceptional skill, adaptability, and a rich range of behaviors, but also harmonic chord selection as well as timing, which makes it challenging for piano-playing cobots [22]. Modern robotic advancements have paved the way for collaborative piano performances involving both humans and robots. For anthropomorphic robots, participating in a piano duet demands rapid decision-making and nuanced sensory feedback. This presents two significant design challenges for a robot aiming to harmoniously accompany human pianists [36]. Firstly, the robot must decide in real-time which expressiveness pattern to adopt, adjusting the music's dynamics, tempo, and articulation accordingly. Secondly, it should continuously calibrate its timing based on the live performance of the human pianist, ensuring a seamless and expressive musical partnership[37], [36]. Even minor deviations in timing can disrupt the flow of the music and therefore cause inharmonious acoustics and communication errors between the human and the robot teammate. For a piano robot, this alignment is made possible through the use of rhythmic oscillatory algorithms that synchronize the entire system of the upper limbs [25], [35].

Regarding collaboration, humans are often regarded as a fast-reacting role due to their significant cognitive capabilities for learning and adaptation. This allows them to supervise the robot partner's superior physical abilities [1]. The interface facilitating human-robot communication is significant in estimating the leading and following roles between the two entities. Effective communication is crucial for high-quality collaboration, i.e., to which extent the robot is able to understand and respond to human non-verbal cues, such as pauses, variations in tempo/rhythm, and keystroke dynamics. In the context of human-robot teamwork in piano playing, it is essential for the robot to establish bidirectional communication with the human operator to achieve a shared understanding of the task and coordinate its movements accordingly. The robot should be capable of interpreting human motion through a reliable communication medium. Both verbal[38], [39] and non-verbal[40], [41] communications have been presented to allow the robot to communicate with the human operator about its internal states and intentions[36], with the purpose of temporal synchronization. Signal timing is crucial to the success of pHRI, particularly in the case of expressive robot companions[42], [43]. Precise temporal synchronization is crucial for minimizing delays, enhancing user experience, and enabling real-time control in collaborative piano performances. It is necessary to build an efficient communication between the human and robot, which facilitates smooth and natural cooperation.

In this paper, we present a co-player designed for human-robot cooperative manipulation in the context of piano playing. Our proposed human-robot collaborative model is designed to empower humans with a responsive partner in expressive piano performances. Figure 1 illustrates the system framework of the human-robot cooperative piano playing. We first proposed a musical improvisation model. The model generates chord accompaniment that aligns with the expressive cues provided by the human pianist. The RNN-based agent draws inspiration from well-established pop songs, delivering chord progressions that resonate with the human teammate. Regarding the physical implementation, we have constructed an anthropomorphic robot arm with an anthropomorphic hand, which is coordinated by the central pattern generator-based controller [44], [45] to achieve motion coordination. To achieve temporal synchronization, we introduced an adaptive controller that facilitates temporal alignment, which is crucial for real-time adaptation to the pianist's performance. Human-robot communication in this context is facilitated through a non-verbal method. The interaction is reciprocal. On the one hand, the robot receives the information generated by the human players. The communication modality here is based on the MIDI signals. The robot's decision-making of chord selection is based on the information generated by human. On the other hand, the human also receives information generated by its robotics teammate. There is an impact of robotic visual and auditory information on human's behavior. Instead of relying on rich proprioception sensing like Electromyography (EMG) measurements [46], this interaction uses the MIDI (Musical Instrument Digital Interface) and RTDE (Real-Time Data Exchange) interfaces, ensuring a bias-free approach. Leveraging MIDI-based parameterization as the communication medium, the robot interacts directly with its human counterpart, seamlessly integrating real-time acoustic inputs and mitigating issues associated with individual bias [12]. We evaluate the performance of our approach in a series of experiments involving adaptive chord accompaniment, and real-time human-robot teamwork in a homophony performance. Experimental results demonstrate the effectiveness of our proposed approach in achieving smooth and seamless cooperative piano playing between human and robot partners. This work contributes to the development of advanced robotic systems that enable human-robot cooperation in creative, expressive and entertainment tasks.

The remainder of the paper is structured as follows: Section II introduces the piano playing platform and the anthropomorphic hand design (end-effector). Section III outlines the development of the chord-prediction agent including the music information retrieval (MIR) algorithm. Section IV explains the adaptive controller enabling temporal synchronization between the robot and its human partner, emphasizing bidirectional communication and information flow. Section V contains the experimental tests and results. Section VI provides a comprehensive discussion of the paper's conclusions.

## II. ROBOTIC PLATFORM

### A. Experimental Setup

The piano cobot consists of three key components: a three-joint motor system, an anthropomorphic robot hand, and the piano environment. The anthropomorphic hand is designed with a hybrid soft-rigid structure, featuring a 3D-printed rigid internal skeleton and a soft silicone skin. The wrist of the anthropomorphic hand is attached to the UR5 robot arm, serving as its end-effector. To conduct the experiments, a Kawai ES8 digital piano is utilized, which supports both the input and output of MIDI signals. For data analysis and real-time communication, Matlab is employed as a software tool. It facilitates the processing and interpretation of experimental data, as well as provides a platform for establishing communication between the cobot and human input through the MIDI device interface. The MIDI signals captured from the piano are inputted into the RNN improvisation program, after which the predicted chord signals are sent to the UR5 control program for physical execution. The runtime for each stage is presented in Table I. This integration facilitates the cobot's ability to promptly and precisely respond to the commands of the human player during piano playing tasks.

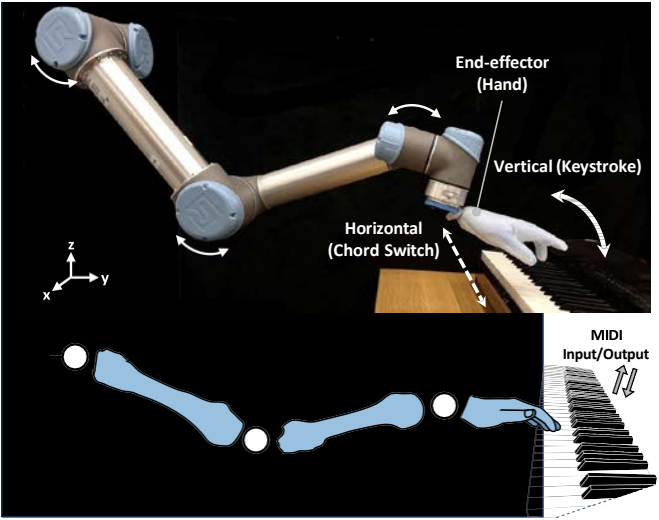

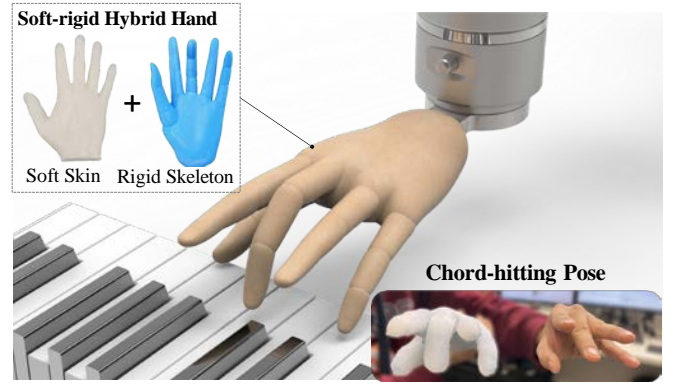

Fig. 2. The robotic setup features a UR5 manipulator equipped with a custom anthropomorphic hand. The earthward (vertical) motion of the arm contributes to the striking of piano keys, while the horizontal motion facilitates chord switching, adding to the overall expressive capabilities of the robot in piano-playing tasks.

Fig. 3. The anthropomorphic hand is fabricated with a hybrid soft-rigid design and preset in a chord-hitting pose.

*1) Keystroke Motor System:* We have utilized the coordinated control of each joint of the UR5 robot manipulator to achieve repetitive and rhythmic piano keystrokes. Our main focus in this research is on the earthward motion, which is a critical aspect of the natural arm swing observed during human piano performances. The earthward motion involves a sequence of coordinated motor movements in which the joints work together. This coordination is particularly important for robots because it enables us to generate the necessary impulse and execute rapid keystrokes on the end-effector within a short time frame. Achieving this level of control and speed is challenging using traditional Tool Central Point (TCP) control and its inverse kinematics. The UR5 robot arm is equipped with an interface that allows for real-time communication and data exchange. This interface enables us to receive the real-time status of the robot arm, including information about its joint pose, velocity, and acceleration. To achieve real-time control over the robot arm, we utilize the RTDE (Real-Time Data Exchange) interface, which facilitates seamless communication between the robot controller and our system. Specifically, we employ the *servoj* function in the Python compiler, which allows us to send precise joint control commands to and receive the sensory data from the UR5 in real-time, enabling us to coordinate its movements accurately during the piano playing task. Figure 2 illustrates the configuration of the robotic setup, which includes a multi-degree-of-freedom UR5 robot arm and a custom anthropomorphic end-effector. This setup controls both the horizontal chord switch along with the piano and the vertical keystrokes.

*2) Soft-rigid Hybrid Hand:* The internal skeleton, responsible for providing structural support, is fabricated using 3D printing technology (Polylactic Acid, PLA). The soft skin, made of silicone (EcoFlex-0020, Smooth-On Inc.), is coated on the external surface of the skeleton. The hybrid soft-rigid design aims to produce the basic anatomic characteristics of a real human hand. Robotic hands without soft skin produce a "rigid-rigid" contact between the finger and the piano key. This rigid contact can lead to relative movement between the fingertip and the piano key, resulting in insufficient key depression. This makes it challenging to control the quality of the keystroke in terms of dynamics and articulation. In contrast, the soft skin enhances friction, creating a "soft-rigid" contact that reduces relative movement between the finger and the piano key, ensuring the quality of each keystroke.

In this study, our focus lies on homophony performance, where the left hand (robot hand) accompanies the melody played by the right hand (users' input). The users are expected to play the melody while the robot generates chord accompaniment. For piano playing, a commonly used chord accompanying technique is triads, i.e., three-note chords. Due to hardware limitations, we simplified the tuning system and focused on several specific chords for piano accompaniment. These chords consist of three major, three minor, and one diminished chord. The selection of these chords was based on considerations of covering various chord types, with the three major and three minor chords being commonly used triads. However, it is worth noting that the diminished B chord is less commonly employed compared to the others. Table II provides details on the chord triads and their corresponding fingering techniques.

| Process | Average runtime (s) |
|---|---|
| MIDI acquisition | 0.0003±0.0007 |
| Signal transfer time | 0.0023±0.0007 |
| RNN prediction + UR 5 command time | 0.0103±0.0026 |
| **Overall control loop** | **0.013±0.003** |

TABLE II
THE CHORDS UTILIZED AND THEIR CORRESPONDING FINGERING TECHNIQUE.

| Type | Name | Piano Notes | | |
|---|---|---|---|---|
| | | Root | Third | Fifth |
| Major Chord | **C** | C | E | G |
| | **F** | F | A | C |
| | **G** | G | B | D |
| Minor Chord | **Dm** | D | F | A |
| | **Am** | A | C | E |
| | **Em** | E | G | B |

To enhance the authenticity and mimicry of human pianists, the anthropomorphic hand's fingers are preconfigured in a chord-playing posture, imitating the natural gesture used to strike major chords on a grand piano (refer to Figure 3). The end-effector of the hand, specifically the little finger, middle finger, and thumb, are positioned on the root note, major or minor third, and perfect fifth notes, respectively. This configuration enables the hand to replicate the chord accompaniment actions performed by human pianists, thereby enhancing the authenticity and human-like qualities of the robot's piano performance.

### B. MIDI-based Musical Modality

The quality of the music played by the piano is a crucial metric, based on which we assess the performance of both the human and robots. We have employed the MIDI protocol, a standardized communication modality that has been widely used to describe and capture various musical instructions such as the pitch, timing, and loudness of individual notes.

In this work, MIDI-based parameterization is employed to facilitate the integration of the robot with the piano-playing task. This enables the translation of musical notes and commands into corresponding robotic actions and gestures. Our primary focus centers on two key components inherent in piano performance: articulation and dynamics. Articulation refers to the timing of keystrokes that ensures the harmonic and coherent coordination of various keystrokes. By controlling the timing of keystrokes, the robot is able to smoothly and accurately reproduce the intended musical flow and rhythms. On the other hand, dynamics concern the strength or velocity applied to the piano keys. For the parameterized musical modality in this work, the loudness is related to the keystroke velocity instead of directly controlled by the force. However, the threshold-crossing velocity is a result of the force exerted on the note by human fingers. Therefore, the piano loudness can be considered as an indirect consequence of the force exerted during keypresses. Our previous work [25] has illustrated how different patterns of expressiveness are achieved by force control on a variable stiffness finger. Regarding the piano setup used in this work, it is noted that a minimum contact force of approximately 20 mN is able to press the piano key downward.

The occurrence of consecutive keystrokes in a repetitive manner can be defined and identified as a threshold-triggering event (Figure 4). When the key displacement reaches a pre-defined threshold level, a MIDI event is triggered, capturing the keystroke velocity ($\omega$) and corresponding time instants. Within a single keystroke, two MIDI events occur: one for key-press (subscript $p$) and one for key-release (subscript $r$). For a single keystroke, approximations of parameters related to articulation and dynamics can be derived:

$$V_{i,p} = r \cdot \omega_{i,p}$$
$$V_{i,r} = r \cdot \omega_{i,r}$$
$$T_{i,hold} = T_{i,p} - T_{i,r}$$

$$(1)$$

where $\omega_i$ and $T_i$ denote the angular velocity and time instant of the $i$th single keystroke. The distance from the axis of revolution to the contact position is denoted as $r$. The velocities during key-press and key-release actions, $V_{i,p}$ and $V_{i,r}$, respectively, are proportional to the keystroke strength. In practical MIDI-based musical applications, the velocity is normalized within the range of 0-127. The duration of the actual keystroke action ($T_{hold}$) plays a crucial role in determining the articulation, representing the length of time each note is played. $T_{idle}$ refers to the gap period when the key is not triggered and is awaiting the next keystroke. For two adjacent keystrokes, the time gap between the key-press instants ($T_{gap}$) depicts how fast the key has been hit in repetitive keystrokes, reflecting the regular interval of sound. It is important to note that the term "tempo" should not be confused with "rhythm", as the tempo governs the articulation control in musical performance, while rhythm relates to the recognizable pattern of note duration. The tempo is measured in beats per minute (BPM) and can be calculated as:

$$T_{i,gap} = T_{i+1,p} - T_{i,p}$$
$$\beta_i = \frac{60}{T_{i,gap}}$$

$$(2)$$

### III. LEARNING-BASED MUSIC ACCOMPANIMENT

To achieve harmonic homophony in the human-robot co-operative piano playing task, effective communication and interpretation between the robot co-player and the human pianist are crucial. The human is responsible for playing the upper voice, while the robot takes on the role of playing the bassline, providing harmonic support to complement the human-played melody. However, a significant challenge for the robot lies in the creative and artistic decision-making process, specifically in selecting the appropriate chords. This requires the robot to make musically informed choices that enhance the overall performance. To address this challenge, we aim to train an agent capable of predicting the ideal chord progression based on the human teammate's performance, taking into account musicality considerations.

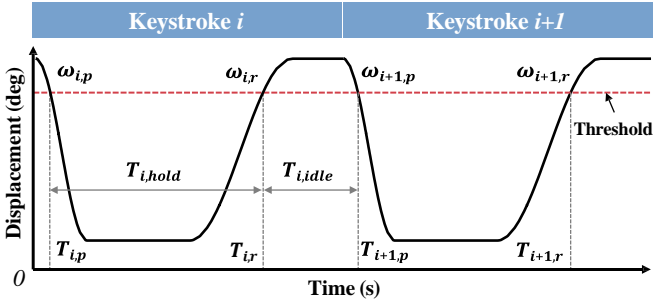

Fig. 4. The MIDI protocol is used as the communication medium for collaborative manipulation.

### A. Dataset Selection

We aim to propose an RNN-based agent that effectively predicts chords in contemporary pop music. To achieve this, we made use of the POP909 dataset [47], an extensive collection of 909 pop songs by 462 musicians, spanning from the 1950s to 2010. Each song in this dataset is represented in a MIDI-based piano format (specifically, the .mid file). Our goal is to understand the essential skills inherent in chord accompaniment within pop songs. A specific chord progression can harmonically support various melodies. This technique has been widely employed in pop songs. For instance, in C major a recurrent chord sequence like "C-G-Am-F" is versatile enough to support the improvisation playing for hundreds of songs, making what might initially be a repetitive, monotonous, and artless melody more dynamic and expressive.

### B. Music Representation and Encoding

During the piano performance, the human player generates a time-series sequence of musical flow through the melody they play. In our approach, we employ a recurrent neural network, specifically a Long Short-Term Memory (LSTM) network [48], to serve as an interactive agent that operates in real-time alongside the human player. The LSTM is structured as a sequence classification model, allowing it to analyze and predict chord progressions based on the input melody. For the future development in the human-robot cooperative framework, addressing large-scale music-related datasets could be facilitated by employing transformer models [49] with parallel computing. To ensure synchronization between the human and the machine, we quantize the music timing into 16th note steps, which represent the shortest duration of notes in our dataset. We assume a 4/4 time signature and adhere to a standard tempo of 90 beats per minute (BPM), facilitating consistent and coordinated performance between the human and the machine.

The RNN model takes in a sequence of 16 notes as its input. The target dataset consists of 7 chord classes, as outlined in Table II, enabling the LSTM network to learn the matching skills between the melody and the corresponding chords. Note that even with octave jumps, these 7 chords still provide significant harmonic support, enabling effective accompaniment. Additionally, in the 16-token sequence, each token can represent any of the 88 piano keys. However, the large size of this input space presents significant computational challenges for the RNN model to effectively learn chord-matching skills. To address this, we utilized the transposition technique to compress them into 12 consecutive indexes, which correspond to the keys within one octave. In scenarios where the human plays multiple notes simultaneously, the note with the longer duration of being held down during that time interval will be selected. Consequently, each note's pitch is encoded using the compressed MIDI index, providing a representation of the melody. We use the variables $\psi_i$ to denote the $i$th human input MIDI pitch number. The compressed MIDI pitch ($x_i$) and the pitch class ($O_i$) can be expressed as:

$$x_i = \psi_i \mod 12 + 1, \tag{3}$$

$$O = \frac{\psi_i}{12} - 1 \tag{4}$$

where mod means calculating the remainder when one number is divided by another, and $\lfloor \cdot \rfloor$ denotes rounding down.

By applying the transposition to all 88 piano keys, we can compress them into 12 categories represented by pitch classes. This compressed representation captures the essential information of the piano keys' pitch and octave. Consequently, the input sequence of n-note melody tokens is denoted as $X_{1:n} = [x_1, x_2, \cdots x_n]$, where $x_i \in \mathbb{N}$, represents the $i$th token encoded with the compressed MIDI pitch number. Let $y$ represent the predicted output. In our RNN network, our goal is to find a model $M$ to approximate the conditional distribution.

$$P(y \mid x_{1:n}) \approx M(y \mid RNN(x_{1:n})) \tag{5}$$

To handle notes with durations longer than a single token (note hold), we repeat the subsequent token until the next distinct note, ensuring that the 16-note bar is fully represented. When no key is pressed at a particular time instant (note rest), the corresponding token is set to 0, indicating the absence of a note. This representation of melody playing is illustrated in Figure 5.

### C. Music Information Retrieval (MIR)

The raw dataset cannot be directly used to feed into the RNN model for training as there is no matching pairs between the melody and the harmonic chord accompaniment. Here we propose a MIDI Chord-Melody Pair Extraction algorithm to extract the essential chord-matching samples that have been widely used for the given dataset, which can be further fed into the RNN-based learning network to train an agent that captures the skills hidden behind contemporary pop music composers. Before the MIR algorithm, the raw music dataset is extracted as .mid file and the MIDI events are ordered in

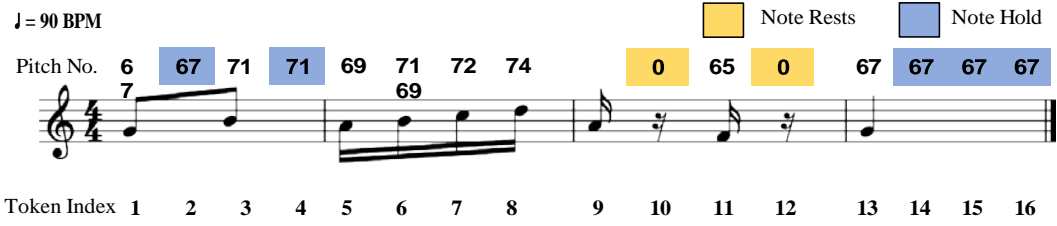

Fig. 5. Tokenization of the MIDI-event sequence. The yellow box represents a note rest (two sixteenth rests here), denoted by the value 0, while the blue box indicates the token holding the note.

---

**Algorithm 1** MIDI Chord-Melody Pair Extraction

**Data:** Black keys set $K_{black}$, Permissible pitch interval set $\Delta P$.

**Result:** Extracted aligned chords and melodies: $\{C, M\}$;

Initialize constants $\delta_{chord}, \delta_{melody}, t_{bar} \rightarrow X$;

**for** *each song* $s \in \mathcal{D}$ **do**

    **for** *each MIDI triplet* $[n_i, n_{i+1}, n_{i+2}]$ *in* $s$ **do**

        $g_1 \leftarrow T_{i+1,p} - T_{i,p}$;

        $g_2 \leftarrow T_{i+2,p} - T_{i,p}$;

        **if** $g_1, g_2 \leq \delta_{chord}$, *and* $\nexists n \in \{n_i, n_{i+1}, n_{i+2}\}$ *s.t.* $n \in K_{black}$, *and triplet pitch intervals* $\in \Delta P$: **then**

            $c \leftarrow [n_i, n_{i+1}, n_{i+2}]$

            **for** *each* $j \in \{1 \dots 16\}$ **do**

                $T_{j,token} \leftarrow T_{i,p} + \frac{(i-1)}{16} \times t_{bar}$;

                **for** *each MIDI event* $k$ *in* $s$ **do**

                    **if** $T_{j,token} + \delta_{melody}$, *and* $T_{j,token} < T_{k,r} \leq \delta_{melody}$ **then**

                        $m_j \leftarrow P_k$;

                    **else**

                        $m_j \leftarrow 0$

                    **end**

                **end**

            **end**

            $m \leftarrow [m_1, m_2, \dots, m_{16}]$

        **end**

    **end**

**end**

**for** *each detected* $c$ *and* $m$ **do**

    Construct the chord-melody pair: $\{C, M\} \leftarrow \{c, m\}$;

**end**

---

a time-series sequence. The songs in the dataset have gone through the aforementioned transposition (raising or lowering the pitch of melody) and the time-stretching [50] to align with the fixed tempo in this work. This is achieved by reallocating the timestamps ($T_i$ in Figure 4) of the MIDI events for all the songs, making each song proportionally faster or slower.

The algorithm presented in Algorithm 1 demonstrates the MIR process. In the initial phase of MIR, we identified timestamps when multiple notes are simultaneously pressed, denoting chords in the dataset. The permissible pitch interval means the difference in pitch number of the three notes in triads, i.e., $\Delta P = \{3, 4, 7\}$. Note that we do not consider chords that fall outside the category of three-finger chords as listed in Table II. Following chord detection, we extracted 16-note melody tokens aligned with each identified chord timestamp. In this study, this chord-melody pair extraction yielded a total of 12,342 $\{C, M\}$ pairs. Our analysis uncovered that the diminished chord *Bdim* was noticeably absent from the compositions within our dataset. This absence can be attributed to the music genre, as pop music typically prioritizes harmonies that are straightforward and immediately appealing to a broad audience. Diminished chords are rarely utilized in pop songs as the sound can be dark and a bit scary, which may not align with the genre's preferences. In comparison to other genres like jazz and classical music, major and minor chords are more frequently employed in pop songs, while diminished chords, known for their dissonant and tense quality, are less commonly used.

It is important to highlight that within these extracted pairs, the same melody input could be associated with different target chord classes. This variability arises from the fact that different musicians may choose different chords based on their individual experiences and skills. Consequently, the chord-matching problem we are dealing with can be viewed as a multi-label classification problem, reflecting the diverse choices made by different musicians. For example, as shown in Figure 6, a single melody sequence can correspond to two or three different chord-accompaniment choices. The observed variations in chord choices among musicians reflect their individual preferences, while also following certain musical rules such as avoiding dissonant diminished B chords, matching the root pitch of the melody, and adhering to a loop-based harmonic structure. Our objective is to train an agent to capture and learn the chord-matching skills within $\{C, M\}$, which is from a wide range of musicians.

### D. Contextual Summarizer

To accomplish the chord prediction task using RNN, we utilize LSTM to capture the hidden state of the RNN and act as a contextual summarizer. We denote the predicted chord across the 7 chord classes, given by $Y = [y_1, \dots, y_7]$. Each $y_j$ corresponds to the probability of the $j$th chord class. The LSTM network predicts the probability distribution of chord options based on the melody tokens. At a particular time step $t$, the forward propagation within the LSTM model can be mathematically expressed as:

$$h_t = \text{LSTM}(w_b h_{t-1} + C_{t-1}) y_t$$

$$Y = \text{softmax}(W_b h_t + b_b)$$

(6)

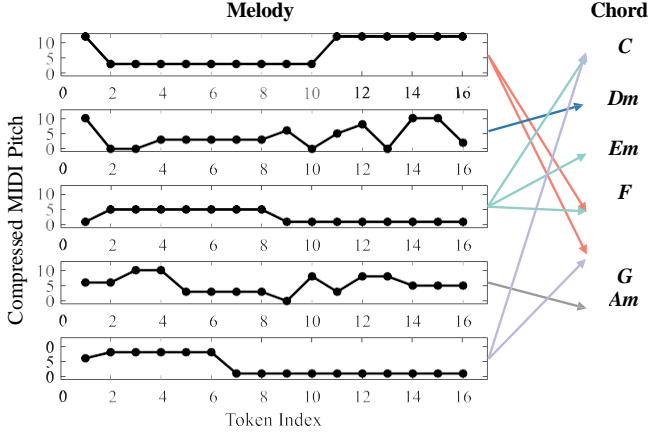

Fig. 6. Multi-label classification for 5 examples of raw data melody sequences.

In the given equation, $x_t$, $h_{t-1}$, and $c_{t-1}$ represent the input, hidden state, and cell state, respectively. The function LSTM($\cdot$) represents the update of the LSTM cell. The weight matrix connecting the hidden state to the output layer is denoted as $W_o$, and $b_o$ represents the bias vector for the output layer of the function softmax($\cdot$) computes the probability distribution over the chorus classes.

The LSTM cell processes the input sequence $X_{1:n}$, updating the hidden state and cell state, and generating outputs at each time step. This process is repeated for all time steps in the input sequence, enabling the network to capture temporal dependencies and make predictions based on the learned context. The final output is then passed through the softmax function to obtain the predicted probability distribution over the 7 chord classes.

Prior to feeding the data into the LSTM model, the dataset is shuffled, and the data is divided into training, validation, and test sets in an 8:1:1 ratio. The network architecture comprises a sequence input layer with normalization, two 80-unit LSTM layers, a fully connected layer with 7 output categories, a softmax layer, and a classification layer. The classification layer selects the class with the highest probability from the output of the softmax layer:

$$\hat{y} = \arg\max(y_j), j = 1, 2...7. \tag{7}$$

where $\hat{y}$ represents the predicted output, and argmax($\cdot$) denotes the function that returns the index with the highest value in the probability vector. To prevent overfitting and improve generalization, we incorporate a dropout layer in the network architecture. During training, the weights of the network are updated using mini-batches of size 256, each containing paired input sequences ($X_{1:16}$, $y$). The learning rate is set as 0.001, with max epochs are 500. We utilize the Adam optimizer, which adjusts the learning rate based on moment estimation, to optimize the model's performance. Since this is a multi-class classification problem, we employ the categorical cross-entropy loss as a measure of training performance. The implementation and training of the LSTM network are conducted using the MATLAB toolbox for deep learning.

### E. Multi-label Classification

In music-related tasks, accurately predicting chords is a complex challenge due to the inherent subjectivity involved. While the confusion matrix is a standard measure of classification performance, it may not be suitable for this task as it does not account for the appropriateness of replacing chords. In the original dataset, particularly within the pop music genre, some chords are interchangeable. For instance, in a given melody, the chord "C" can be substituted with "F" while still providing harmonic support for the original melody. However, it is worth noting that this prediction accuracy is an average across all chord labels. When it comes to the true chord labels (ground truth), the confusion matrix only accounts for true positives. In reality, for a target chord, there might be one or more other chords also appropriate replacements. To reflect the model's real prediction accuracy, we developed a compensation model for the prediction accuracy, taking into consideration the analysis of chord replaceability within the original dataset after applying Music Information Retrieval (MIR) techniques.

We first compute the confidence level of using one chord to replace another by summarizing the pairs after MIR. Algorithm 2 outlines the detailed procedure for computing the directional probability. First for all the chord-melody pairs,

Here, $N(\cdot)$ denotes the number of occurrences of one label $N$ is the pairs by enumerating pairs with each chord, maximum $N$ ← 1 chord label (RC). Next, the algorithm distribute the remaining $N$ ← $i$ in chord-melody $N$ ← 1, the algorithm distribute the remaining chords in the given dataset $c$ as designating the remaining chords as replaceable chords (RC). The ultimate directional probability, denoted as $P_{\mu\nu}$, signifies the confidence level of

---

**Algorithm 2** Directional Probability of Chord Replacement

**Result:** Confidence level; $P_{TC_\mu \leftarrow RC_\nu}$, $\mu \leq 7$ and $\nu \leq 7$.

**Data:** Extracted chord-melody pair from MIR's $C$, $M_i$; classification pairs, and set $x$ here not one-class.

**for** each normalized chord-estimated do

  $c_{label}^i \leftarrow c_1^i, c_2^i, c_3^i \cdots \varsigma^i$, $j \geqslant 2$.

  **for** $k \in 1, 2 \cdots j$ **do**

    **if** $k = \arg\max N$ $c^{i,k}$ **then**

      $TC^i \leftarrow c_k^i$

    **else**

      $RC^i \leftarrow c_k^i$

    **end**

  **end**

**end**

**for** $\mu \in 1, 2 \cdots 7$ **do**

  **for** $v \in 1, 2 \cdots 7, v = \mu$ **do**

    $P_{TC_\mu \leftarrow RC_v} = \dfrac{\sum_i N(RC^i)}{\sum_i N\ TC_v^i}$.

  **end**

**end**

using chord $C_v$ to substitute chord $C_\mu$ within this particular music genre.

Originally the prediction accuracy of the confusion matrix is calculated by summing the diagonal elements to get the total number of correct predictions and summing all the elements in the confusion matrix to get the total number of predictions, where the dividing leads to the accuracy:

$$n = \frac{\sum_{7} C_{ii}}{\sum_{i=1}^{7} \sum_{j=1}^{7} C_{ij}} \qquad (8)$$

where $C_{ii}$ indicates the count of instances where the true chord is correctly predicted while $C_{ij}$ represents the count of instances where the true chord $c_i$ is predicted as other chords $c_j$. For the prediction accuracy considering the chord replacement, we assign the confidence level, and the refined accuracy is predicted as:

$$n' = \frac{\sum_{7} C_{ii}}{\sum_{i=1}^{7} \sum_{j=1}^{7} C_{ij} - \sum \lambda_{\mu\nu} \frac{\sum_{7} \sum_{7} CR_{\mu\nu}}{CR_{\mu\nu}}} \qquad (9)$$

where $RC_{\mu\nu}$ denotes the count of instances where the target chord $c_\mu$ is feasible to be replaced with chord $c_\nu$. The coefficient $\lambda$ serves as an indicator and $\psi$ denotes the threshold:

$$\lambda_{\mu\nu} = \begin{cases} 1 & : P_{\mu\nu} \geq \psi. \\ 0 & : P_{\mu\nu} < \psi. \end{cases} \qquad (10)$$

$$\psi = \min(P_{\mu\nu}) + \xi \cdot (\max(P_{\mu\nu}) - \min(P_{\mu\nu})) \qquad (11)$$

threshold of the chord replacement coefficient to adjust the $\xi \in [0, 1]$ here, the strength coefficient is 0.19.

### F. Sliding-window Arrangement

Both the human and the robot contribute to shaping the decision-making process for the upcoming musical bar. The trained model addresses the challenge of chord selection. However, musical expressiveness is influenced by various factors, including the chosen chord, pitch, duration of each chord, and the manner in which multiple chords are connected (progression and articulation). In order to react to the melody playing speed, the speed of chord playing is determined by counting the number of pitch variations within a given musical bar. We have mapped this count to a function that represents the duration of chord presses. As illustrated in Algorithm 3, for the defined tokenized 16-note bar, for the $i$ th ($i \in [1, 16]$) note, if the current MIDI pitch $X_i$ is different from the previous one $X_{i-1}$, the total pitch variation, denoted as $\tau$, will add 1. $\tau \in N \cap [1, 16]$, the function $D(\tau)$ represents the number of chord presses within a single musical bar and is defined in Equation 13.

The timing of chord presses is realized by evenly distributing the initial time instant among the 16 tokens. The alignment of the melody tokens and the chord timing is illustrated in Figure 7. The heavy beats of the chord strokes are evenly distributed at the beginning of each subdivision, spacing them

---

**Algorithm 3** One-bar Pitch Variation

**Data:** Melody signals from human input $X_{1:n}$, $n \in N$; Tempo $\beta$; Time signature $X/Y$.

**Result:** The number of pitch variations within the bar: $\tau$.

Initialize constants: $\tau = 1$; $\delta_{melody}$; $t_{bar} \leftarrow \frac{60}{\beta} \times X$;

**for** $j = 1$ *to* 16 **do**
  $T_{j,token} \leftarrow T_{i,p} + \frac{(j-1)}{16} \times t_{bar}$;
**end**

**for** $k = 1$ *to* 16 **do**
  **if** $T_{i,token} \geq T_X^{k} - \delta_{melody}$ *and* $T_{k,token} < T_X{}^k{}_{,r} + \delta_{melody}$ **then**
    $\tilde{X}_k \leftarrow X_k$;
  **else**
    $\tilde{X}_k \leftarrow 0$;
  **end**
**end**

**for** $i = 1$ *to* 15 **do**
  **if** $P_{\tilde{X}_i} \quad x_{i-1}$ *and* $P_{\tilde{X}_i} \neq 0$ **then**
    $\tau \leftarrow \beta + 1$;
  **end**
**end**

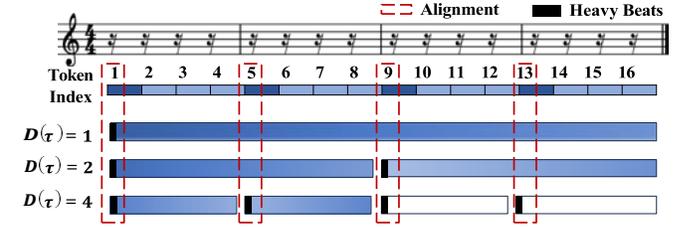

Fig. 7. The chord timing alongside the melody token sequences, with color decay corresponding to acoustic strength degradation.

evenly throughout the sequence. For example, if $D(\tau) = 4$, they are aligned with the 1st, 5th, 9th, and 13th tokens.

The co-player's focus is twofold: improvising the bassline based on the human's input and determining how long the chord should be matched. Only when all 16 tokens of the melody have been fed into the RNN model can the model give a chord prediction. This means that the robot cannot improvise on the current bar at the same time, resulting in a lag. Due to this inherent melody-to-chord lag, aligning the predicted chord with the human input sequence poses a challenge. We employ a sliding window technique, as illustrated in Figure 8. This technique aims to produce the $p$th chord accompaniment ($y^{(p)}$) and determine the number of consecutive keystrokes (CK) ($D^{(p)}$) based on the last melody $X^{(p)}_{1:16}$ and last chord ($y^{(p-1)}$). Figure 8 depicts that the chord accompaniment window is slid by one bar. This means that the chord predicted from the previous bar is used to provide harmonic support for the current homophony window, allowing for aligned accompaniment. The process can be expressed as:

$$y^{(p)} = M\left(y/RNN\left(X^{(p-1)}, y^{(p-1)}\right)\right) \qquad (12)$$

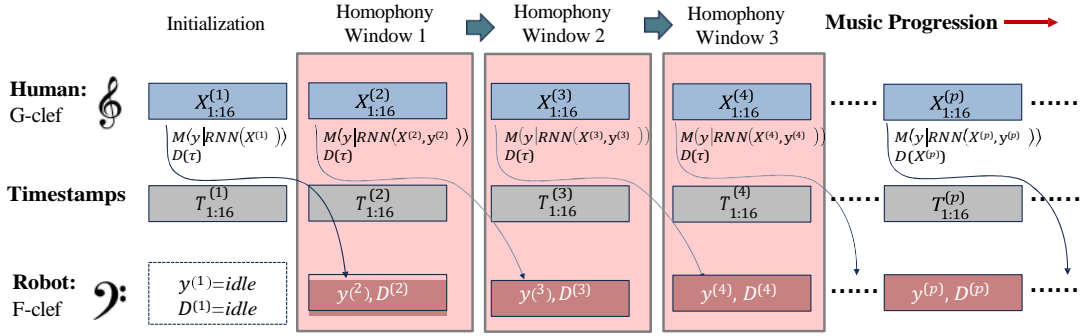

Fig. 8. The sliding-window technique for melody-chord alignment.

$$D^{(p)} = 1 + \left\lceil \frac{\tau(X_{1:16}^{(p-1)})}{9} \right\rceil + 2 \cdot \left\lceil \frac{\tau(X_{1:16}^{(p-1)})}{13} \right\rceil \quad (13)$$

## IV. COOPERATIVE PLAYING SYSTEM

Achieving precise temporal synchronization between the robotic system and human input is pivotal for the seamless execution of harmonious and coordinated musical performances. In this study, a challenge lies in the meticulous determination of the temporal alignment of chord-hitting and chord-switching actions, encompassing the arrangement of both the vertical and horizontal movements of the end-effector. This intricate task is further complicated by the inherent physical constraints governing the robotic manipulator's capabilities.

To tackle this challenge, we propose a hierarchical control architecture. The entire problem is conceptualized as a setpoint tracking task. The low-level control provides basic motion control for the horizontal chord switching and coordinates the anthropomorphic upper-limb joints for vertical chord strokes. The supervisory control is responsible for trajectory planning and synchronization management. A model predictive control (MPC) is utilized to orchestrate the sub-controllers including an RNN agent, trajectory planner and CPG controller.

### A. End-effector Trajectory

*1) Vertical Motion:* For a single-chord keystroke, the vertical motion of the UR5 is controlled as a coordinated arm swing behavior. The coordination of the upper-limb joints is based on our previous work [35]. Regarding the vertical motion, three joints of the UR5 robot receive the output from a refined central pattern generator (CPG) and enable the hand to produce the earthward keystrokes. It outputs an impulsive trajectory (as depicted in Figure 9), which outperforms the sine-wave curve-based trajectory in terms of prompt keystroke. The core of the controller is based on the Matsuoka neural rhythmic oscillator [45].

$$T_r \frac{dx_i}{dt} + x_i = -\sum_{j=1}^{n} a_{ij} y_j + s_i - b f_i \quad (14)$$

$$T_a \frac{df_i}{dt} + f_i = y_i \quad (15)$$

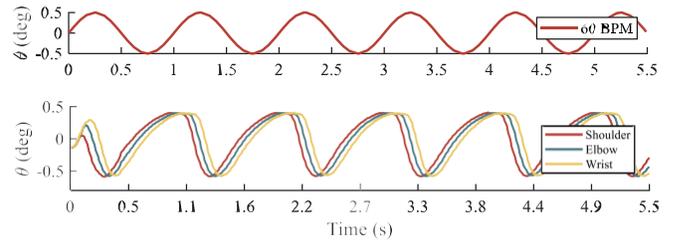

Fig. 9. Comparison of the trajectory between a sine wave curve (upper) and the CPG-based trajectory (bottom) for impulsive keystrokes.

$$y_i = g(x_i) \quad (16)$$

In Equations 14, 15 and 16, we have $i, j \in \mathbb{N}$ with $j \neq i$,

$$y = \begin{cases} x & : x \geq 0 \\ 0 & : x < 0 \end{cases} \quad (17)$$

The variables $x_i$ and $f_i$ correspond to the membrane potential and fatigue characteristics, respectively, as observed in biological systems. The parameters $T_r$ and $T_a$ play a pivotal role in governing the response time of $x_i$ and $f_i$, influencing their dynamic characteristics. Additionally, $s_i$ signifies the excitatory tonic input, while $a_{ij}$ and $b$ represent the magnitudes of reciprocal inhibition and self-inhibition.

*2) Horizontal Motion:* Horizontal motion serves the purpose of positioning the end-effector at different chord locations, allowing adequate time for the vertical motion to complete the chord-hitting action. Prior to execution, the UR5 robot undergoes a calibration process to align its physical position with the benchmark note on the piano. Human leverages the dexterity and agility inherent in human hand movements, enabling pre-planned path execution and swift chord transitions during piano performance. In the case of robots, the goal is to minimize the time consumption for chord-switching, but constraints arise due to hardware limitations. These constraints encompass the physical span between chords and the duration of each chord press, necessitating adjustments in the horizontal motion speed of the end-effector, denoted as $V$ (mm/s).

For a given tempo of $\beta$ beats per minute (BPM) and a time signature of $X/Y$, where $X, Y \in \mathbb{N}$. We have the duration of each chord press, $T_{bar}^{(i)}(D(\tau))$, as denoted in Equation 13 and 18. Let $d$ represent the width of each piano white key and $h$ denote the number of piano notes traversed by the end-effector during chord switches ($h \in \mathbb{N}$), we need to adjust the chord-switching motion to ensure that an appropriate amount of time, $\zeta$, is allocated within each chord press.

$$T_{bar}^{(i)} = \frac{60 \cdot X}{\beta} \tag{18}$$

$$\zeta \cdot \underbrace{v \cdot T_{bar}^{(i)}}_{D(\tau)} = h \cdot d \tag{19}$$

Here, $\zeta$ represents the time portion within the last chord keystroke used for moving horizontally for the given consecutive keystroke (CK) condition. Typically, we have $\zeta \le \frac{1}{2}$ to ensure that the keypress of this keystroke has been completed, even though some of the time for key release has been used for compensating the chord switching motion. In this study, the control of each joint of the UR5 robot manipulator is orchestrated to accomplish horizontal chord switching, the extreme condition is that the robot arm needs to traverse the allowance, i.e., $D(\tau) = 4$ and $h = 6$). Due to the hardware limitation (CK is not agile/fast enough), the required motion speed $v$ in the horizontal direction for this transition and the time allowance $\zeta$ should be dynamically adjusted following Equation 19. The controller needs to strike a balance between fulfilling the task rapidly and staying within the limitations of the UR5 manipulator[51]. The introduction of $\zeta$ addresses situations where the maximum speed of the TCP motion may not be sufficient for chord switching. In such cases, some of the time allocated for chord keystrokes is used to compensate for horizontal motion. This compensatory mechanism accounts for environmental perturbations and guarantee the trajectory tracking performance during the execution of the UR5 manipulator task.

### B. Human-robot Cooperation Architecture

*1) Trajectory Planner:* The management of the end-effector's movement is framed as a trajectory-tracking problem. As illustrated in Section III-F, the sliding window technique reveals a consistent one-bar gap between the ongoing melodic segment and the corresponding chord accompaniment. Leveraging the proposed Recurrent Neural Network (RNN) model, the system can improvise the tokenized melodic input provided by human users and formulate decisions regarding the accompanying chord progression. Once the impending chord progression is ascertained, a desired trajectory is generated from the current chord position to the subsequent predicted chord. Iteratively, it enables the robot to deliver real-time musical accompaniment. Figure 10 provides an illustrative depiction of a planned trajectory with the chord and CK information following the sequential chords as outlined in Table III.

Within a musical progression scenario involving human input, the RNN-based music accompaniment model in Equations 12 and 13 performs dual tasks by predicting the appropriate chord to be used and specifying the consecutive keystroke (CK) conditions. Simultaneously, the algorithm generates a planned trajectory. This trajectory can be rapidly updated and created for the next few steps by utilizing both the current end-effector position and the path predicted by the RNN agent. The UR5 robotic arm is then instructed to follow this pre-planned trajectory. As depicted by the nodes in Figure 10, the end-effector executes horizontal movements to reach the desired positions corresponding to different chords, followed by performing CPG-based vertical keystrokes. The objective of approximating the desired trajectory is achieved by providing control inputs that guarantee the end-effector's arrival at the specified locations at precise timestamps.

| Bar | 1 | 2 | 3 | 4 | 5 | 6 | 7 | ... |
|---|---|---|---|---|---|---|---|---|
| Chord | C | Em | G | Dm | Em | F | Am | ... |
| CK | 2 | 4 | 1 | 2 | 2 | 4 | 2 | ... |

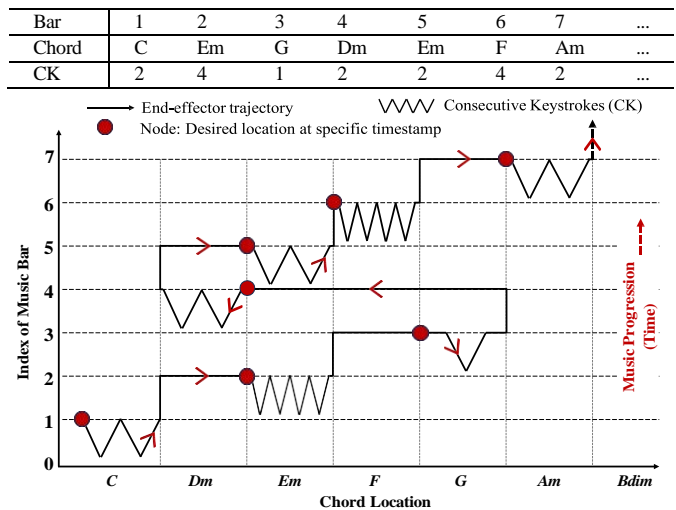

Fig. 10. Schematics of a planned trajectory of the end-effector along with the chord progression corresponding to Table III.

*2) Motion Synchronization:* In order to create harmonious music accompaniment with precise timing, it is essential for the robot to move to the intended location promptly. As the distance between the current chord and the predicted next chord is continuously changing, the planned trajectory undergoes constant updates. This necessitates our controller to dynamically execute the end-effector. We have developed a hierarchical controller, as depicted in Figure 11. Both the human and the robotic teammate have mechanical contact with the piano (task environment). Music-related information is captured and encoded as input, which is then fed into a trained RNN agent. The RNN model improvises the tokenized musical bar, determining which chords to use for harmonic accompaniment and the number of consecutive keystrokes (CK) for the upcoming bar. Once the decision-making for the musical accompaniment is complete, the trajectory planner generates the desired path for the end-effector. The robot's controller is responsible for following this reference trajectory. The low-level component of the hierarchical controller handles CPG-based keystroke actions and horizontal motion control for chord-switching behavior. Meanwhile, the high-level control involves model predictive control (MPC) for adjusting the horizontal motion speed and time allocation, denoted as $v$ and $\zeta$, with the aim to timely control the robot once the RNN model has improvised the MIDI signals such that the temporal synchronization is met. Here, $Q_x$, $Q_y$ and $Q_z$ represent position of the end-effector, also known as the Tool Center Point (TCP). The control variables are denoted as $V = (v, \zeta)$, where $v = [u_1, u_2, u_3, u_4, u_5, u_6]$ and each $u_i$ corresponds to the joint velocities of the 6 joints of the UR5 robotic arm. For the dynamics of the robot system, we have the following model:

$$\dot{x}_k = f_{UR}(x_k, V)$$
$$x_{k+1} = x_k + \dot{x} \cdot \Delta t \tag{20}$$

The function $f_{UR}(\cdot)$ encompasses the transformation operations, which include both the forward kinematics transformation and the Jacobian matrix. This Jacobian matrix establishes a connection between the velocity of the Tool Center Point (TCP) and the joint velocities [51]. The second equation within Equation 20 represents the updates of joint angle and end-effector coordinate. The MPC aims to minimize a cost function that quantifies the discrepancy between the current state $x_k$ and the desired reference trajectory $x_k^*$. The cost function is defined as follows:

$$J(x, v) = \omega_1 \sum_{k=0}^{N} \|x_k - x^*_k\|^2 + \omega_2 \sum_{k=0}^{N} \|v_k\|^2 \tag{21}$$

where $N$ represents the prediction horizon, and $\omega$ stands for the weighting coefficient. The second term in the cost function reflects the penalty associated with control effort. Regarding constraints, there is a limit on the joint velocity of motion, and the chord-switching behavior must align with the motion control, as specified in Equation 19. Additionally, since we have confined chord selection to the chords listed in Table II, there are constraints on the spatial coordinates of the end-effector. Therefore, we have the following constraints:

$$0 \leq v \leq v_{\max}$$
$$x_{\min} \leq x \leq x_{\max} \tag{22}$$

We have the dynamic optimizer working as:

$$\zeta = \frac{1}{2^R}, \quad R \in \mathbb{N} \tag{23}$$

Minimize $J(x, v)$

subject to:
$$\dot{x}_k = f_{UR}(x_k, v), \qquad \forall k = 0, 1, \ldots, N,$$
$$0 \leq v_k \leq v_{\max}, \qquad \forall k = 0, 1, \ldots, N,$$
$$x_{\min} \leq x_k \leq x_{\max}, \qquad \forall k = 0, 1, \ldots, N. \tag{24}$$
$$\zeta = \frac{1}{2^{R}}, \qquad R \in \mathbb{N}, \quad \forall k = 0, 1, \ldots, N.$$

In this work, we consider two control inputs. The first one is the horizontal motion speed, and the second one is the time allocation, as indicated in Equation 19. The second control variable is conditional. When it involves moving short distances, the robot is capable of reaching a specific position the chord keystroke behavior, i.e., $u_i = 1$. However, for longer-distance chord switching, such as transitioning from chord G to chord F, the robot's agility is insufficient, even when operating at maximum speed. Therefore, we need to reallocate some of the time from the vertical keystroke motion for chord switching. In this study, we use a binary partitioning technique to determine the time allowance for chord switching, ensuring that the previous chord is held for the desired duration while reserving sufficient time for chord transitions. The time allocation parameter $\zeta$ is based on Equation 23.

The entire problem is formulated as an RNN-based Model Predictive Control (MPC) system, which follows an iterative optimization approach. During each sampling time interval $k$, the current state is measured, and an optimal input vector is determined by minimizing a cost function while adhering to various constraints. MPC offers the advantage of considering multiple constraints, which is crucial due to the numerous physical limitations in both communication and the motion of the UR5 robot. The binary partitioning of the minimal chord time duration acts as a time allocation for chord transitions in the continuous flow of music progression. This mechanism is implemented to rectify any discrepancies in note timing, ensuring that both the human and robot strike the notes simultaneously.

### C. Reciprocal Interaction

In Figure 11, effective communication is crucial for collaborative piano playing, where the interaction is not one-sided; the human influences the robot, and vice versa. Both the robot and the human rely on non-verbal cues for communication, which encompass modalities like MIDI, visual feedback, and audio feedback. This communication involves the exchange of information in both directions. However, this bidirectional information flow could lead to issues such as undesirable leading or lagging during the collaboration. To address this challenge, both the musical improvisation model and the cooperative controller are regularly refreshed and calibrated. As detailed in Section III-F, the music progression is processed in continuous sequential musical bars, with each bar's length determined by the musical notation. The musical improvisation is adjusted at each bar interval, ensuring that even if the human plays faster or slower than the ideal tempo, the robot will still strike the chord at the correct timestamp. This approach prevents unwanted self-induced leading or lagging, thereby maintaining the quality of the music throughout the performance. A similar framework focusing on the human-robot co-playing of musical instruments has been proposed [52]. This paper presents a controller designed to enable a robotic player to synchronize with a human flutist, where the angle of the flute is tracked and utilized as input to generate robot control signals. In contrast, our study focuses on piano instrumentation. We leverage a MIDI-based protocol modality, which remains robust even in the presence of disturbances from human or environmental factors. The captured non-verbal cues are processed using a learning-based model instead of relying on hard-coded programming to facilitate music accompaniment. In the realm of information theory, let us consider two time series variables $S_H$ (representing human performance) and $S_R$ (representing the source signals from robot actions). The question is whether $S_R$ "Granger-causes" [53] $S_H$, indicating that the past values of $S_R$ contain information that improves the ability to predict the future values of $S_H$ when compared to predicting $S_H$ solely based on its own past values. Our primary objective is to understand how both the robot and the human can effectively utilize non-verbal cues to enhance their piano-playing performance during collaboration. To address this, we have introduced the following transfer entropy ($TE$) to quantify information flow from robot actions ($S_R$) to human actions ($S_H$):

$$TE(S_R \rightarrow S_H) = H(S_H | S_{H-\varphi}) - H(S_H | S_{H-\varphi}, S_{R-\varphi})$$

(25)

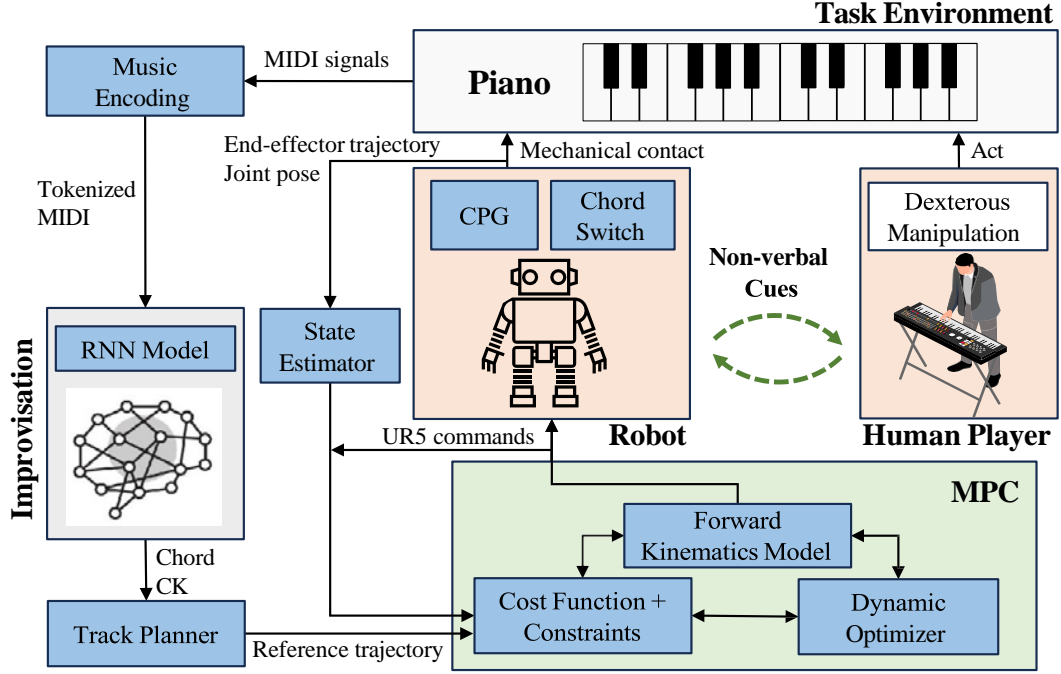

Fig. 11. Overall controller architecture of the cooperative piano playing system.

In this context, $H(S_H | S_{H-\varphi})$ signifies the entropy related to predicting the human's performance based solely on the human's past actions. $H(S_H | S_{H-\varphi}, S_{R-\varphi})$ denotes the entropy associated with predicting the human's performance by considering the historical actions of both the human and the robot. To evaluate the quality of cooperation, we analyze the synchronization of keystroke timings between the robot and the human by introducing the entropy for assessing the time deviations. The variables are the timestamps and keystrokes of the human and the robot. The deviation is the time difference between the human's action and the reference action. It could be positive (if the human is ahead) or negative (if the human is lagging). We can have the entropy of the time deviation distribution by using the Shannon entropy [54] to measure the uncertainty in the time deviations.

$$H(\Delta) = - \sum_{i=1}^{N} P(\Delta_i) \cdot \log(P(\Delta_i))$$

(26)

Where $P(\Delta_i)$ is the probability of observing a particular time deviation, which is based on the frequency of these events in the entire dataset. In addition, we introduce two metrics for analyzing the nuanced temporal cooperation, namely the Mean Absolute Error (MAE) and Sum of Absolute Errors (SAE). Given two sequences of keystroke timestamps $T^H$ and $T^R$ of length $N$, where $T_i^H$ and $T_i^R$ are the data points at $i$th keystroke by human and robot, respectively. The MAE and SAE are calculated as:

$$\text{MAE} = \frac{1}{N} \sum_{i=1}^{N} |T_i^H - T_i^R|$$

(27)

$$\text{SAE} = \sum_{i=1}^{N} |T^{H_i} - T^{R_i}|$$

(28)

In our physical implementation (Figure 12), we conducted a feedback manipulation experiment to validate the directional information flow between the robot and the human. This experiment involved piano playing while deliberately blocking one or more feedback modalities. The current considerations of the robot-to-human impact are visual and auditory signals. However, another consideration can be the expressiveness. For example, the consecutive keystroke (CK) condition in one musical bar. Different tempo and CK conditions can reflect the robot's expressive patterns, which can also influence the human's performance. To assess the information flow from the robot to the human, we conducted four groups of experiments during subjective tests for human-robot cooperative piano playing. We established four experimental conditions with different degrees of visual and audio feedback: no visual and no audio (NV-NA), no visual with audio (NV-A), with visual and no audio (V-NA), and with visual and with audio (V-A). To block visual feedback, participants wore an eye mask, while noise-canceling headsets were employed to obstruct audio feedback from the robot to the human, including both piano sound and motor noise.

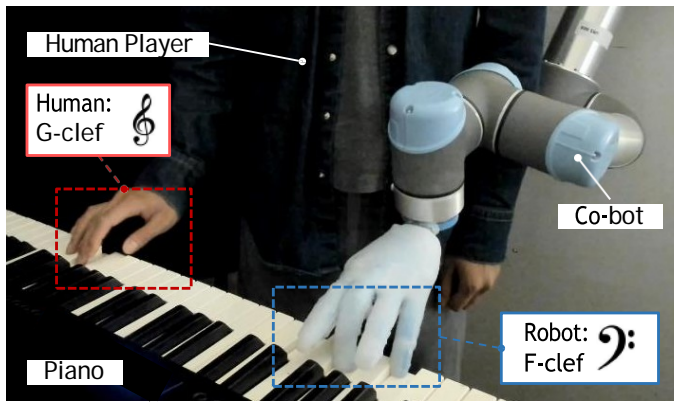

Fig. 12. The experimental setup for the human-robot cooperative piano playing, where the human is in charge of the upper voice while the robot is providing the lower voice with chord accompaniment.

TABLE IV
DIRECTIONAL CONFIDENCE LEVEL OF REPLACEABLE CHORDS

|    | Am | C | Dm | Em | F | G |
|----|----|----|----|----|----|----|
| **Am** | 1.00 | <u>0.34</u> | 0.00 | <u>0.06</u> | <u>0.23</u> | <u>0.12</u> |
| **C** | <u>0.44</u> | 1.00 | <u>0.11</u> | <u>0.28</u> | <u>0.10</u> | <u>0.20</u> |
| **Dm** | 0.00 | <u>0.08</u> | 1.00 | 0.00 | <u>0.54</u> | 0.00 |
| **Em** | 0.00 | <u>0.14</u> | 0.02 | 1.00 | 0.00 | <u>0.31</u> |
| **F** | <u>0.23</u> | <u>0.08</u> | <u>0.50</u> | <u>0.04</u> | 1.00 | 0.00 |
| **G** | <u>0.05</u> | <u>0.04</u> | <u>0.03</u> | <u>0.22</u> | <u>0.03</u> | 1.00 |

## V. RESULTS

### A. Chord Accompaniment Prediction

*1) Chord Prediction Accuracy:* In pop music, there are conditions where more than one chord may be suitable for supporting the same melody sequence, and these chords can be interchangeable due to the subjectivity of musicians. To provide a comprehensive assessment, we consider the prediction accuracy while taking the appropriately replaceable chords into account.

Using the original training dataset and following Algorithm 2, we have created Table IV, which displays the directional confidence levels of replaceable chords. In this table, the rows of chords represent the target chords, while the columns denote the substitute chords. This structured representation offers valuable insights into the confidence levels associated with directional chord replacements. It captures the appropriateness of replaceable chords, summarizing chord replacement conditions captured from the original pop music dataset. By applying the refined prediction accuracy formulation described in Equation 9 and 10, we can determine the threshold for each chord replacement and subsequently update the list of appropriate replaceable chords. These interchangeable chords are indicated by underlined chord classes in Table IV, reflecting a more precise understanding of when chord substitutions are suitable in pop music accompaniment.

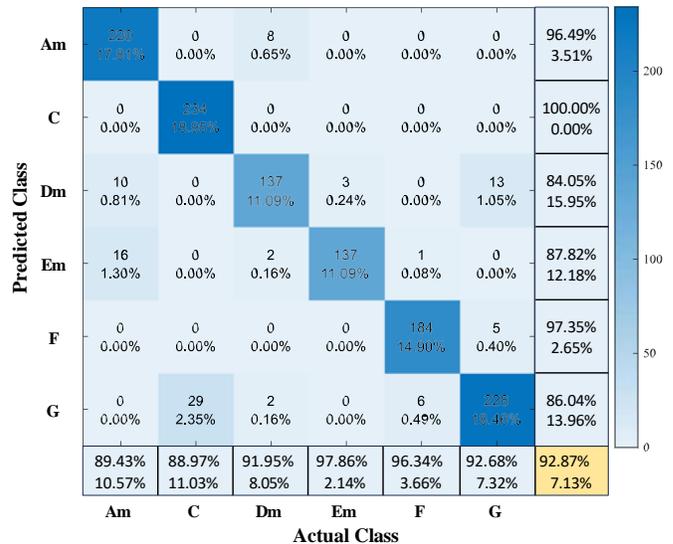

Fig. 13. The confusion matrix of the learning-based chord prediction. Where the horizontal labels denote the target chords, and the vertical axis denote the predicted classes. The cell denotes the number of predictions and its percentage using the test dataset.

Using this information, we have the confusion matrix that considers the subjectivity inherent in chord substitutions. Figure 13 illustrates the refined confusion matrix, offering a more precise and comprehensive evaluation of the model's performance by considering the intricacies of chord replacement conditions in pop music. Note that for a target chord class, the prediction of substitute target chords is also considered correct for the calculation of prediction accuracy. For instance, when the target chord is Dm, the RNN model predicting either chord C or chord F is considered correct. Other cells in the matrix represent chords that are unsuitable for substitution. Based on the analysis of appropriate chord replacement, the proposed RNN model is capable of achieving an overall prediction accuracy of 92.87%.

In real-world scenarios of human-robot cooperative duet performance, various factors such as human mood, physiological condition, and environmental influences can affect the music produced. To validate the accuracy of chord prediction in such real-world experiments, two human participants are instructed to collaborate with the cobot in performing the same duet piece (*Auld Lang Syne*). Each participant performs independently without prior knowledge of the other. Figure 14 illustrates the chord prediction performance of the learning-based improvisation in this real-world setting. In Figure 14a, the piano-roll performance of the cooperative duet playing is depicted, showing that both players provide similar melodic input in the upper voice, while the robot generates the same chord predictions as the music accompaniment (Figure 14b). The observation of different consecutive keystroke (CK) conditions for the two chord accompaniments (occurring at timestamps 4.3s for player 1 and 9.4s for player 2) in the lower voice is due to variations in the human's melodic input from the previous musical bar (based on Equation 13.

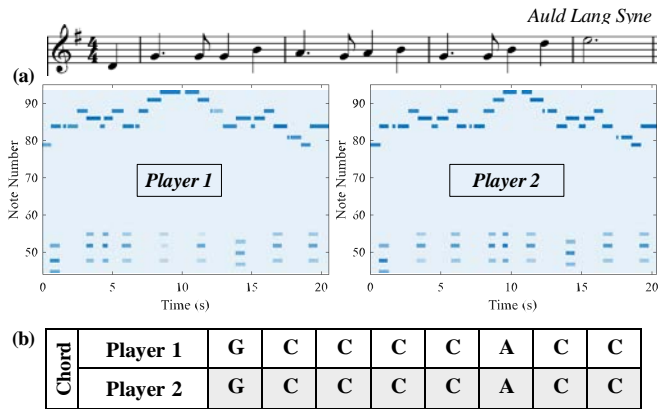

Fig. 14. The chord predictions in real-world human-robot cooperative piano playing. (a) The piano-roll illustration of the duet performance. (b) The results of the chord predictions for two players.

### B. Human-Robot Cooperation

*1) Temporal Synchronization:* In accordance with the controller architecture depicted in Figure 11, the robot demonstrates its capability to deliver real-time musical accompaniment for a human performer through the application of an RNN-based improvisation model. We utilized the developed system in an actual human-robot collaborative piano-playing task. Our participant was an experienced pianist (male, 28 years old) with over 20 years of expertise. The pianist was instructed to use his right hand to perform the melody of a song (*Castle in the Sky*) twice consecutively, with the first rendition at a regular tempo and the second at an accelerated pace. Our analysis focused on evaluating the robot's real-time accompaniment effectiveness.

Figure 15 shows the outcomes of this collaborative performance, consistent with the content provided in the Supporting Video. Figure 15a depicts the piano-roll illustration of the piano-playing performance, showing all the keystroke information generated by both the human pianist and the robotic collaborator. In this representation, the x-axis corresponds to the time window, while the y-axis indicates the specific piano keys that were pressed. The figure encompasses all the music-related information including the pitch of the note, the timestamps (both key press and key release) and duration of each keystroke, and the strength of the keystroke. In this human-robot duet piano performance, it can be seen that the robot promptly responds to the human's musical input, providing harmonious chords that complement the human-generated melody, which validates that the cobot with an RNN-based agent is able to effectively provide the music support. The onset of the human player's initial keystroke marks the beginning of the performance. Analysis of the timestamp data for both human and robot actions reveals an absence of robot activity during the first four beats, even though the human pianist had already started playing. This time lag aligns with the one-bar delay inherent in the robot's real-time accompaniment, as indicated by the sliding-window technique in Figure 8. Figure 15b illustrates the trajectory of six joints. Notably, Joint 2, Joint 3 and Joint 4 of the UR5 robot play a pivotal role in generating synchronized, rhythmic piano keystrokes in the vertical axis. Meanwhile, the control over all six joints of the UR5 robot is utilized to facilitate chord-switching behavior along the horizontal axis.

Regarding tempo transitions, we can observe instances where the robot smoothly shifts from a 2-CK (two consecutive keystrokes) pattern to a 4-CK pattern. This occurs notably at time points 51.0 seconds, 93.73 seconds, and 101.75 seconds, with the robot seamlessly transitioning to a 4-CK pattern within a single musical bar. These transitions highlight the robot's ability to tune its chord accompaniment rhythmic patterns, as denoted in Equation 13. Additionally, the robot exhibits a dynamic speed control capability, enabling it to adapt its movement speed to match its human counterpart. For instance, at 43.1 seconds, the human pianist suddenly accelerates, resulting in an overall faster tempo. Consequently, during the second performance, there is a notable increase in the occurrence of both 2-CK and 4-CK patterns compared to the slower initial rendition. This observation underscores the effectiveness of the robot's response in terms of synchronizing with the musician's musical articulation.

To further investigate the temporal alignment of human-robot collaboration, we conducted experiments on four additional musical compositions. Throughout the musical sequences, we recorded the timestamps corresponding to the onset of the primary heavy beats within each musical bar. As depicted in Figure 16, we captured the time gaps (TGs) between the heavy beats produced by the human and those from the robot. These differentials were computed by subtracting the robot's timestamp from the human's. At each time step within a musical bar, a TG value can be derived. A positive TG indicates that the human is leading the robot, while a negative TG suggests that the human is behind the robot. Throughout the musical progression, TG represents the dynamic lead-lag relationship between the human and the robot. We interpret these time-varying TGs as a function of the musical bars. Figure 17 presents the TGs illustrating the temporal alignment in human-robot collaboration across the four music pieces.

Aside from the time gap, we introduce the concept of progression time gap as the temporal difference between successive heavy beats in the musical performance. The instantaneous phase comparison between the human and robot provides comprehensive insights into the robot's temporal alignment with the human collaborator. We employ synchronization indices (SI) to represent the phase locking value [55], quantifying the degree of phase synchrony. An SI of 1 indicates perfect synchronization, which means that the robot achieves a perfect phase locking with its human teammate in terms of musical progression. In contrast, an SI of 0 indicates no synchronization. The SI can be computed using the following equation.

$$SI = \frac{1}{N} \sum_{n=1}^{N} e^{i(\phi_H[n] - \phi_R[n])} \qquad (29)$$

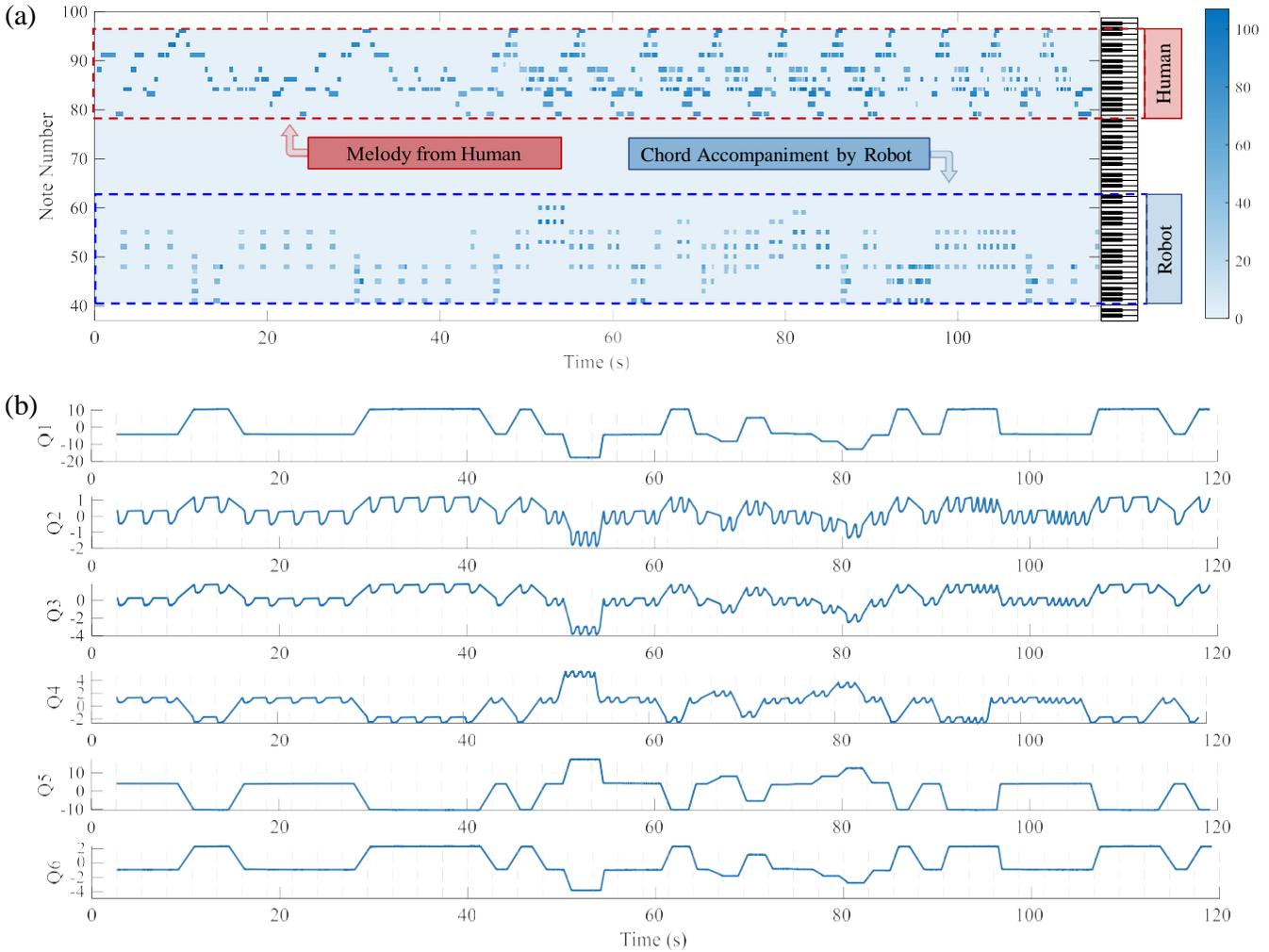

Fig. 15. Human-robot cooperation for expressive piano playing. (a) The piano-roll illustration for the collaborative task, featuring the upper voice representing human input and the lower voice indicating chord accompaniment generated by the robot. The color map represents keystroke strength on a scale of 0-127. (b) The trajectory (degree) of the 6 joint angles, where joints 2 (shoulder), 3 (elbow), and 4 (wrist) coordinate the arm movements for arm-swing keystroke generation.

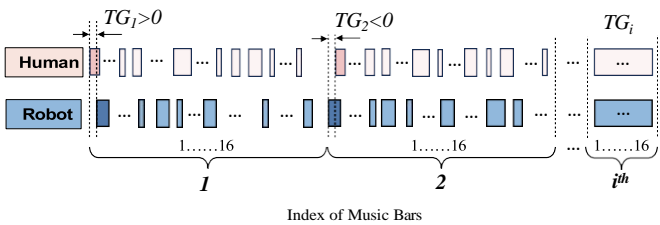

Fig. 16. Illustration of the time gap (TG) along with the music progression.

where $N$ represents the number of taps, and $\phi_H$ and $\phi_R$ represent the phases of each dyad member (human and robot), respectively. Consequently, the Table V lists the average time gap for the human-robot cooperation and the synchronization index. It can be seen that for all the four additional music pieces, the average time gaps between the human and the robot remain below 0.06s. Furthermore, in each musical piece, both the human and the robot demonstrate high levels of synchronization.

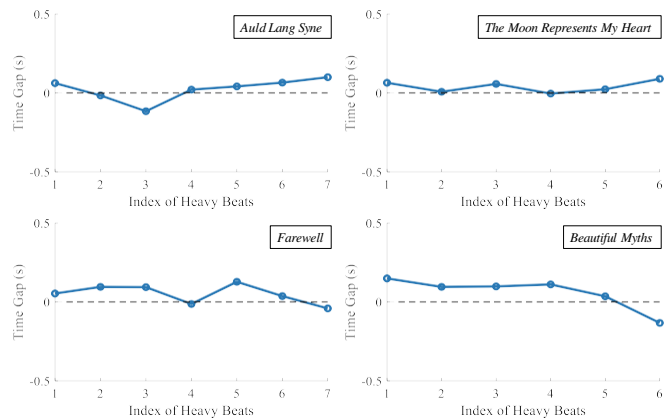

Fig. 17. The time gaps of human-robot synchronization on multiple music pieces.

2) *Control of the End-effector:* In the framework of the proposed Model Predictive Control (MPC) based control ar-

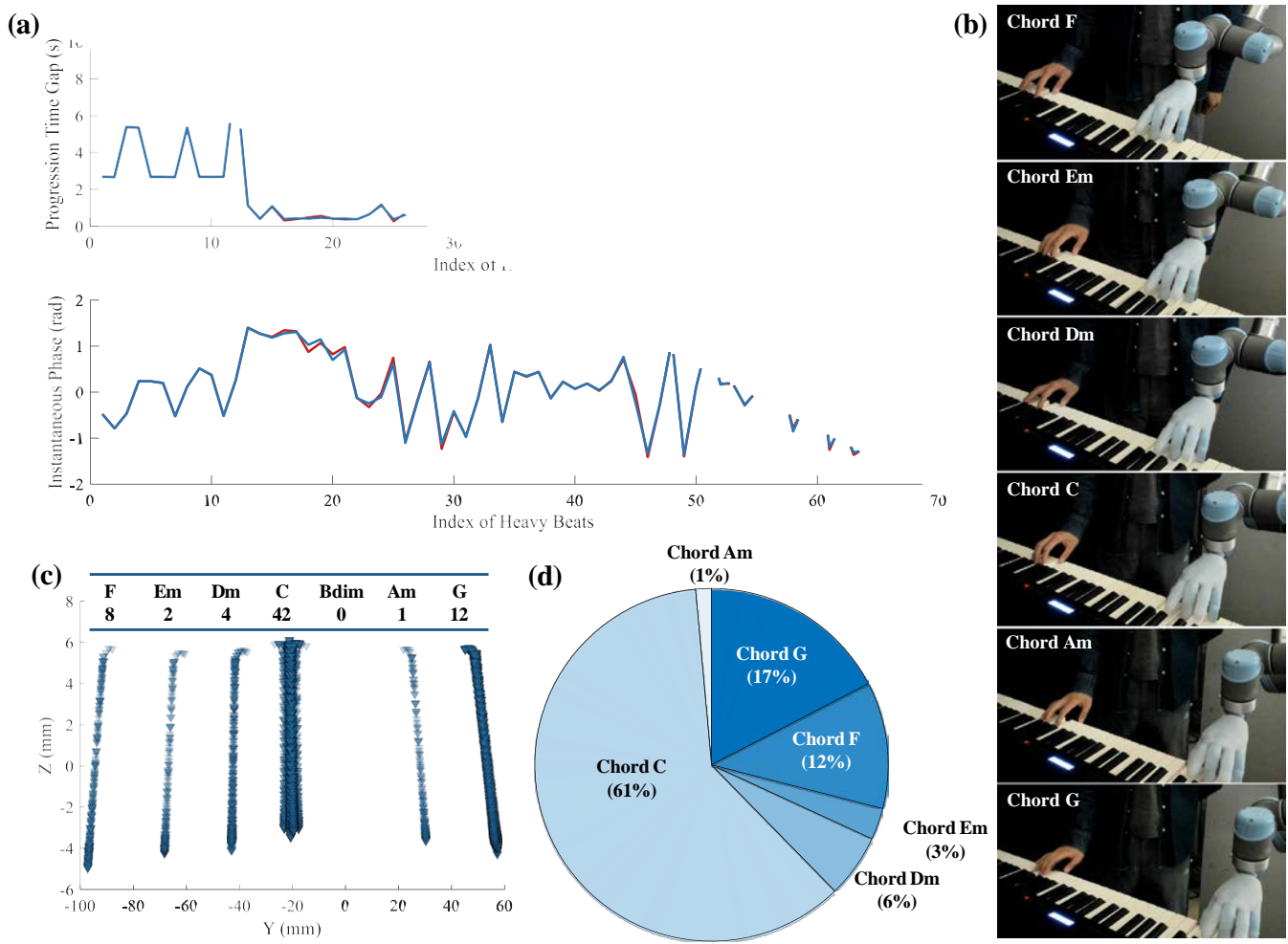

Fig. 18. The analysis of human-robot synchrony and summary of chord accompaniment. (a) The progression time gaps of heavy beats and the instantaneous phases within the music progression. (b) Illustration of the spatial orientation of the end-effector for various chord keystrokes. (c) The trajectory of the chord visits, where the transperance reflects the frequency of visits. (d) Summary of the percentage of each chord utilized for the duet piano performance.

TABLE V
THE PERFORMANCE OF TEMPORAL SYNCHRONIZATION BETWEEN THE HUMAN AND THE ROBOT.

|  | Auld Lang Syne | The Moon Represents My Heart | Farewell | Beautiful Myths |
|---|---|---|---|---|
| Average TG | 0.0225 | 0.0396 | 0.0504 | 0.0594 |
| SI | 0.9997 | 0.9998 | 0.9978 | 0.9998 |

chitecture, we conducted an examination of the phase synchronization between the human and the robot. As illustrated in Figure 18a, we present the progression time gaps of heavy beats alongside the music progression. This metric signifies the temporal difference between successive heavy beats in the performance, reflecting the coherence of the musical flow.

As depicted in Figure 18a, the progression time gaps of the robot player closely approximate those of the human player. The phases of each player align well, resulting in a high phase locking. After the analysis, the synchronization index (SI) value is determined to be 0.9987, indicating a strong phase locking between the human and the robot throughout the music progression. Figure 18b shows the realistic spatial location and orientation of the robot's hand when pressing different chords. In the context of our collaborative duet piano performance, Figure 18c highlights the number of chords improvised by the Recurrent Neural Network (RNN) and their frequency of use on the YZ plane. It can be seen that, for a given human melody input, the robot primarily provides chord support with major chords, such as chord C (utilized 42 times) and chord G (utilized 12 times), while the usage of minor chords, such as chord Em and chord Dm, is less frequent. This observation aligns with the characteristics of pop music, known for its infectious hooks and a preference for a more positive and upbeat sound. The pie-chart in Figure 18d further delineates the percentage distribution of real-time chord accompaniment generated by the robot during the cooperative piano playing.

## C. Information Flow during Coooperation

In order to validate that genuine cooperation occurs between the robot and its human counterpart during piano playing, our objective is to examine the bidirectional communication within the human-robot system. As Figure 1 has illustrated, the human and robot engage in mutual communication through non-verbal signals. Regarding the information transfer from the human to the robot, the robot interprets MIDI-based cues to determine the appropriate chord and its arrangement. Conversely, in the communication from the robot to the human, a combination of visual and audio cues serves as the communication modality. The human partner utilizes their sensory organs, specifically their eyes and ears, to monitor, interpret, and even anticipate the intentions of their robotic counterpart.

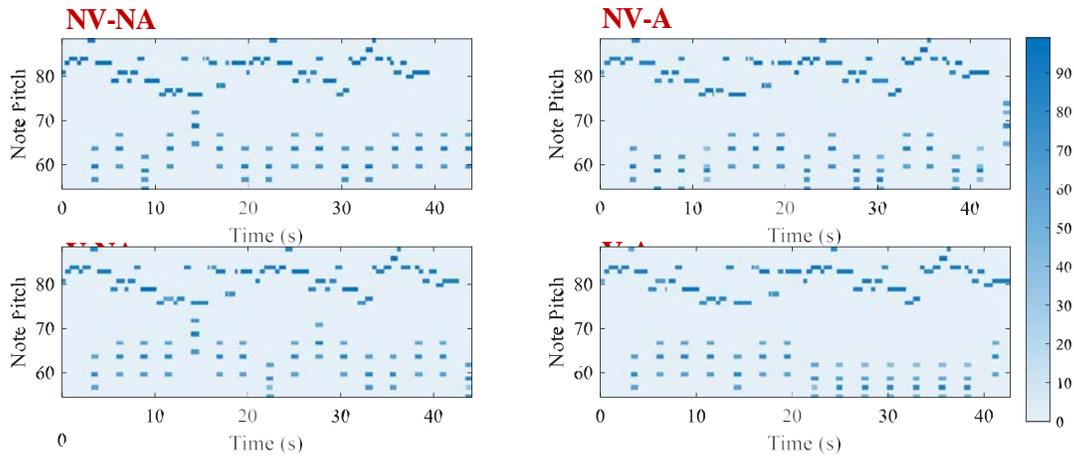

Fig. 19. Schematics of the piano roll during the subjective test for human-robot cooperative piano playing. The four groups denote the conditions with controlled feedback, i.e., no visual no audio (NV-NA), no visual with audio (NV-A), with visual no audio (V-NA), and with visual with audio (V-A). The color bar shows the MIDI velocity, where the keystroke velocity is normalized to 0-127.

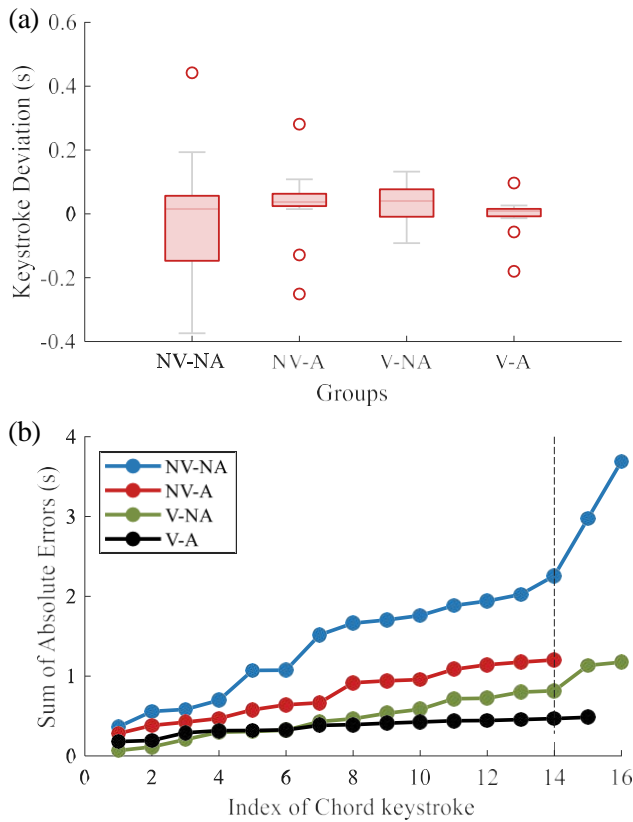

Fig. 20. The accuracy of the cooperative keystrokes. (a) The box-and-whisker plot of the keystroke deviations between the human and the robot for chord matching. (b) Accumulate cooperation error (denoted by the sum of absolute errors, SAE) for four groups.

Our primary goal is to validate the information flow from the robot to the human performer. To accomplish this, we conducted a controlled experiment where we systematically blocked either visual or audio feedback from the robot, leading to varying degrees of influence on the human player's performance. Under these controlled conditions, the participant was instructed to perform the same melody in each of the four defined scenarios, with each participant repeating the performance five times, and the results were averaged for analysis. The piano-roll illustrations in Figure 19 depict these scenarios, i.e., no visual, no audio (NV-NA), no visual with audio (NV-A), with visual, no audio (V-NA), and with visual with audio (V-A). It can be seen that in the NV-NA condition, the human player exhibited a progressively faster tempo in the upper-voice region, with the entire melody performance lasting less than 40 seconds. This acceleration is due to the lack of effective external feedback, preventing the player from calibrating their keystrokes with a reference beat. In the other three conditions, the cooperative performance remained relatively consistent, with slight variations in the number of chord keystrokes (16, 14, 15, and 14, respectively), as even minor deviations in the timing of melody keystrokes influenced the result of RNN model's chord improvisation, despite participants being instructed to perform the same melody.

Furthermore, an assessment of the temporal synchronization between the human and the robot provides valuable insights into the quality of their cooperation. This analysis involves the examination of all chord occurrence time instances, with a focus on chord accompaniment timestamps to determine the human input timing deviation. The comparison of keystroke deviations across the four cases is depicted in Figure 20a, where the red line within each box represents the median value of the dataset. The bottom and top edges of the box represent the first quartile (Q1) and third quartile (Q3). The ends of the whiskers extend to the minimum and maximum values within 1.5 times the interquartile range. Data points beyond the whiskers (red circles) are considered outliers. The box-and-whisker plot reveals significant differences for the four scenarios. The NV-NA case exhibits substantial variance, indicating a poor level of cooperation, resulting in inharmonious acoustic output. In contrast, with the introduction of either visual or audio feedback, the quality of cooperation improves, albeit with a slightly larger variance in the V-NA case compared to the NV-A case. This suggests that, in the context of a single-sense scenario, humans may rely slightly more on auditory cues. The most robust performance is observed in the V-A case, characterized by the lowest standard variance. Furthermore, we quantified the Sum of Absolute Errors (SAE) and Mean Absolute Errors (MAE), as illustrated in Figure 20b and Table VI. It can be seen that SAE values increase as the musical progression unfolds, except for the V-A case, which reaches a near-saturation state, signifying excellent cooperation. The lowest MAE is observed in the V-A case, registering at only 0.0323 seconds, in contrast to the single-sense cases (NV-A and V-NA) with MAE values of 0.086 and 0.0734 seconds, respectively. The NV-NA case exhibits poor cooperation, as reflected in the temporal deviation of 0.2307 seconds, significantly impacting the harmony of the human-robot duet piano performance. Based on the Shannon entropy formulation provided in Equation 26, we have computed the entropy for all four scenarios, and the results are depicted in Table VI. It can be seen that for the three feedback-based scenarios, the entropy demonstrates a reduction compared to the NV-NA case, which initially holds 3.2028 bits of information. According to Equation 25, it can be seen that the introduction of visual feedback, audio feedback, or both leads to significant reductions in Shannon entropy, which quantifies the level of uncertainty. Specifically, the entropy drops by 0.79 bits, 0.25 bits, and 1.14 bits, respectively. This observation validates the notion that the robot's performance "Granger-causes" the human performance, indicating an information flow from the robot to the human. This information exchange serves to enhance the collaborative nature of piano playing.

TABLE VI
COOPERATION PERFORMANCE IN TERMS OF MEAN ABSOLUTE ERRORS (MAE), SUM OF ABSOLUTE ERRORS (SAE) AND SHANNON ENTROPY.

| | NV-NA | NV-A | V-NA | V-A |
|---|---|---|---|---|
| MAE(s) | 0.2307 | 0.0860 | 0.0734 | 0.0323 |
| SAE(s) | 2.2583 | 1.2033 | 0.8163 | 0.4680 |
| Entropy(bit) | 3.2028 | 2.4138 | 2.9528 | 2.0635 |

## VI. CONCLUSION AND DISCUSSION

This paper introduces a theoretical framework of human-robot collaboration in the domain of musical and entertainment robots, with the aim of enhancing expressive and creative performances. Our study introduces a music improvisation model that leverages a recurrent neural network to generate chord harmonies that seamlessly align with the expressive cues provided by human pianists. Inspired by popular music, this agent produces chord progressions that resonate with contemporary musical styles. Achieving temporal synchronization is a pivotal aspect, and the robot's real-time adaptation to the human pianist's performance is essential for a seamless collaboration, where we fulfill it by presenting a model predictive controller (MPC) that facilitates accurate timing and temporal alignment.

Our approach to human-robot communication relies on MIDI, which is a non-verbal form of communication modality, fostering a conducive environment for the robot to seamlessly integrate real-time acoustic inputs, promoting more effective collaboration with human musicians. Through a series of experiments, we have demonstrated the efficacy of our approach in achieving smooth and harmonious cooperative piano playing between human and robot partners. During the experiment, we also assessed the dynamics of information flow by employing Shannon entropy as a valuable tool. Finally, the Granger causality relationship between the robot and the human is shown. In future iterations of the system, prioritizing human safety is imperative. One potential approach involves integrating distance sensors or force sensors onto the UR5 robot arm. This would enable the arm to promptly stop its trajectory if a human is detected in its path, thereby preventing potential injuries.

Our experimental results demonstrate the system's capacity to efficiently facilitate human-robot collaboration in the realm of piano playing. The developed platform could serve as a valuable tool for piano beginners and individuals with disabilities, offering them opportunities for expressive musical implementation. Our primary quality assessment focuses on quantitatively measuring keystroke dynamics and articulation, such as timestamps and gaps. Evaluating whether a listener "enjoys" the music involves psychological and subjective analysis. With enhanced dexterity and agility in the robot's hand, the robot can produce sounds that better satisfy listeners' psychological and subjective expectations. The feedback from human participants in this study indicates that the human play- ers felt fatigue and easy to collaborate with the co-bot during duet piano performances. This is partly due to the robot's design, which reduces the control complexity and allows for higher tolerance in managing keystroke quality. Consequently, human players experience less fatigue and pressure, as the robot effectively compensates for errors or mistakes they might make. This success underscores the potential of our approach and its ability to serve as a valuable foundation for further advancements in the field of human-robot interaction and creative expression. To the best of the authors' knowledge, this work marks the first instance of designing and implementing a collaborative robot expressly engineered to work with humans in the artistic pursuit of piano playing. Future work can be developed by incorporating more chords other than the chord triads shown in this work to enhance the diversity of the robot- based music accompaniment. Enhancing the dexterity of the end-effector is crucial to enable it to adapt its position and orientation effectively for pressing a broader range of chords. Moreover, as the complexity of chord accompaniment patterns increases, there will be a greater number of potential melody-chord combinations. Developing an improvisation algorithm capable of managing intensive computational resources while preserving chord prediction accuracy is essential for future endeavors in this field.

## ACKNOWLEDGMENT

This project has received funding from the European Union's Horizon 2020 research and innovation program under the Marie Sklodowska-Curie grant agreement No. 860108.

## ETHICS STATEMENT

This study involving human participants was conducted in accordance with the principles embodied in the Declaration of Helsinki and local statutory requirements. The research protocol was approved by the Research Ethics Committee of the Department of Engineering, University of Cambridge. All participants provided informed consent prior to participating in the study. The experiment was designed to be non-invasive, and no harm or discomfort was caused to the participants during the data collection process. Confidentiality and anonymity of the participants were strictly maintained, and any identifying information was handled securely. The study was conducted with full respect for the rights, dignity, and welfare of the participants. Data handling and storage were carried out in compliance with applicable data protection regulations. This study contributes to scientific knowledge and understanding in the field of human-robot cooperative piano playing, while upholding the ethical standards and principles of research involving human participants.